\pdfoutput=1

\documentclass[11pt]{article}

\usepackage{ACL2023}

\usepackage{times}
\usepackage{latexsym}
\usepackage{graphicx}
\usepackage{multirow}
\usepackage{multicol}

\usepackage[T1]{fontenc}

\usepackage[utf8]{inputenc}

\usepackage{microtype}

\usepackage{inconsolata}
\usepackage{booktabs}

\title{YAYI-UIE: A Chat-Enhanced Instruction Tuning Framework for Universal Information Extraction}

\author{Xinglin Xiao$^{1}$, Yijie Wang$^{1}$, Nan Xu\thanks{$~~$Corresponding author}$^{~1,2}$, Yuqi Wang$^{1,2}$, Hanxuan Yang$^{2}$, Minzheng Wang$^{2}$,\\ \textbf{Yin Luo}$^{1,2}$, \textbf{Lei Wang}$^{1,2}$, \textbf{Wenji Mao}$^{2}$ \and \textbf{Daniel Zeng}$^{\ast 2}$ \\
        $^{1}$Beijing Wenge Technology Co., Ltd. \\  $^{2}$MAIS, Institute of Automation, Chinese Academy of Sciences \\
         \texttt{\{xinglin.xiao,yijie.wang,nan.xu,yuqi.wang,yin.luo,lei.wang\}@wenge.com}    \\     \texttt{\{yanghanxuan2020,wangminzheng2023,wenji.mao,dajun.zeng\}@ia.ac.cn}  }

\begin{document}
\maketitle
\begin{abstract}
The difficulty of the information extraction task lies in dealing with the task-specific label schemas and heterogeneous data structures. Recent work has proposed methods based on large language models to uniformly model different information extraction tasks. However, these existing methods are deficient in their information extraction capabilities for Chinese languages other than English. In this paper, we propose an end-to-end chat-enhanced instruction tuning framework for universal information extraction (YAYI-UIE), which supports both Chinese and English. Specifically, we utilize dialogue data and information extraction data to enhance the information extraction performance jointly. Experimental results show that our proposed framework achieves state-of-the-art performance on Chinese datasets while also achieving comparable performance on English datasets under both supervised settings and zero-shot settings.   
\end{abstract}

\begin{figure*}[ht]
\centering
    \includegraphics[width=6.2in]{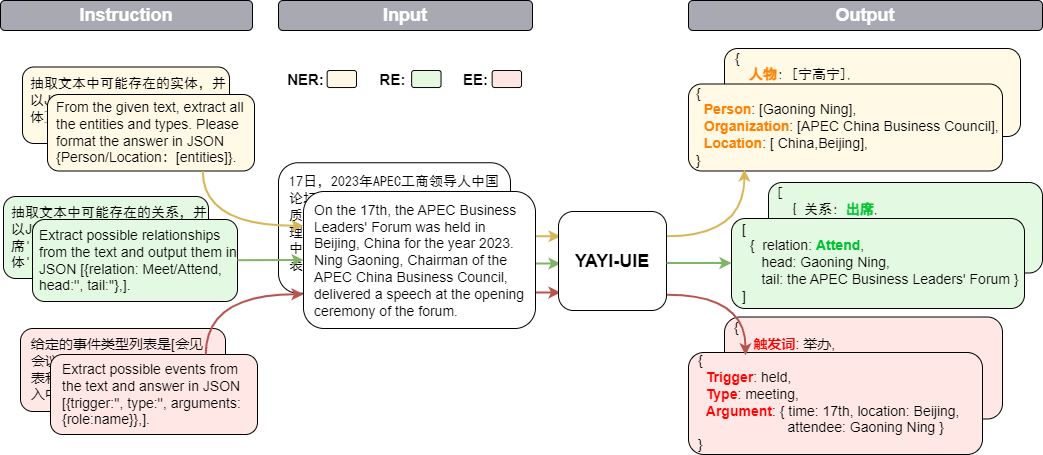}
  \caption{Examples of our chat-enhanced instruction tuning framework for universal information extraction.}
  \label{fig-schema}
\end{figure*}

\section{Introduction}
Information extraction (IE) aims to extract structured information from unstructured text automatically \cite{grishman2019twenty}. Depending on the extracted objects, the IE tasks can be categorized into multiple sub-tasks, including named entity recognition (NER), relation extraction (RE), event extraction (EE), and so on. The traditional IE methods mostly develop isolated datasets and models for each task, schema and domain, which greatly hinders the practical applications of the IE tasks \cite{mengge-etal-2020-coarse,wang-etal-2021-cleve,qin-etal-2021-erica,wang-etal-2022-miner}. 

Recently, large language models (LLMs) have demonstrated tremendous capabilities in solving a variety of natural language tasks and are equipped with strong generalization abilities. Therefore, Lu et al. \cite{lu2022unified} first introduced the concept of universal IE to uniformly model various IE tasks. They also proposed a large-scale pre-trained universal IE model called UIE. However, UIE still requires model fine-tuning for different downstream tasks, which leads to its poor performance on unseen data. Lou et al. \cite{lou2023universal} proposed USM and designed three unified token-linking operations to decouple various IE tasks, but its training and inference processes suffer from inefficiency. Wang et al. \cite{wang2023instructuie} developed an end-to-end unified information extraction framework InstructUIE based on instruction tuning, which utilizes descriptive instructions to enable LLMs to understand different IE tasks. Nevertheless, these existing methods are deficient in their IE capabilities for Chinese languages other than English.

In this paper, we propose YAYI-UIE, an end-to-end chat-enhanced instruction tuning framework for universal information extraction that supports both Chinese and English. Our framework consists of the following two instruction-tuning steps. The first step involves utilizing dialogue data to fine-tune a base LLM for obtaining a chat model with common understanding abilities. In the second step, we focus on enhancing the chat model's performance in IE tasks. To achieve this, we construct the largest and most comprehensive Chinese IE benchmark dataset and combined it with the existing English benchmark. The universal IE model is obtained by instruction-tuning the chat model using this combined dataset. 



\begin{itemize}
  \item  We propose an end-to-end instruction tuning framework YAYI-UIE for universal information extraction that supports both Chinese and English, which leverages dialogue data and information extraction data to enhance the information extraction performance jointly. 
  \item  We construct the most comprehensive Chinese instruction tuning benchmark for universal information extraction, which consists of 16 datasets from various domains.
  \item  The experimental results demonstrate that our YAYI-UIE achieves the SOTA performance in both supervised and zero-shot settings for Chinese, while also displaying remarkable proficiency in English. 
\end{itemize}

\begin{figure*}[t]
  \begin{center}
    \includegraphics[width=5.7in]{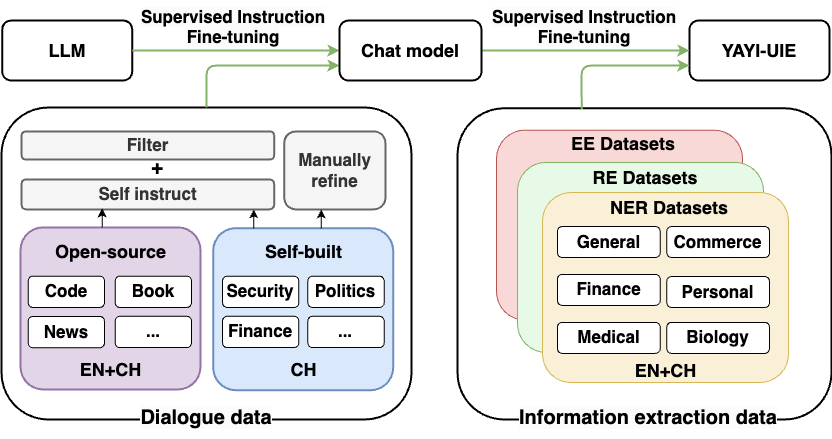}
  \end{center}
  \caption{Overview of our chat-enhanced instruction tuning framework for universal information extraction.}
  \label{fig-overview}
\end{figure*}

\section{Methodology}
In this section, we describe the proposed chat-enhanced instruction tuning framework for universal information extraction (YAYI-UIE). We start with the task schema and the inference procedure of our universal information extraction framework. Then we introduce the design of the two-step instruction tuning, including instruction tuning for chat and information extraction respectively. 

Figure \ref{fig-schema} gives examples of our text-to-text generation framework for universal information extraction to illustrate the task schema and the inference procedure. To uniformly model the IE tasks, including NER, RE and EE, we formalize these tasks by the following task schema:
\begin{equation}
   \textbf{Ouput} = \textit{YAYI-UIE} ~(\textbf{Instruction}, \textbf{Input})
\end{equation}
where the detailed descriptions of the properties in the schema are as follows:
\begin{itemize}
    \item \textbf{Instruction} is a natural language text sequence that includes three elements: task type, task option, and output format. It consists of a description of the task type 
    to specify the task; a description of the task option to restrict the range of the labels in the output; and a description of the desired format of the output. 
    \item \textbf{Input} is a textual instance of the IE tasks that is fed to the large language model along with the instruction, and the model generates the output based on the constraints provided by the given instruction. 
    \item \textbf{Output} is a sentence that represents the structured information extracted from the input text. Specifically, our YAYI-UIE chooses JSON as the output format for all the IE tasks.
\end{itemize}

On this basis, we design a two-step instruction tuning for universal information extraction. As shown in Figure \ref{fig-overview}, we first fine-tune a pre-trained LLM on the dialogue instruction corpus to enhance the instruction-following ability. Following that is the instruction tuning for information extraction, which aims to better constrain the model to generate the desired structured results for the IE tasks.

\subsection{Instruction Tuning for Chat}
To enhance the model's understanding of open-world languages and improve the performance of instruction fine-tuning in fully supervised and zero-shot settings, intuitively, the dialogue data in real life is a good fit for strengthening the understanding of human language instructions. In the first step of the proposed two-step instruction tuning, we use open-source dialogue data with instructions and a self-constructed corpus to train a chat-enhanced language model to facilitate instruction tuning for multiple information extraction tasks. 

\paragraph{Dialogue Data}
During the data acquisition, to align the model for following human instructions better, we first perform general instruction tuning to train a chat-enhanced model using dialogue corpus in both English and Chinese. The corpus is sourced from the general internet webpages and public datasets, including high-quality data such as news articles, encyclopedic contents, books, codes, etc. In addition to the open-source datasets, we also leverage some field-specified data in the domains of finance, politics, and security. These self-built data, with a large portion in Chinese, include press conference records, company identification, and sensitive boundary recognition, etc.

For data processing, the corpus is constructed based on the self-instruct framework \cite{wang2022self}, formatted as tuples of instruction, input and output. Specifically, we iteratively perform instruction tuning using the instances generated by our model. At each iteration, the distribution of the generated data is revised using a filtering step, where the meaningless, incomplete, sensitive, or duplicate samples are rejected. In addition, for the field-specified data, we further manually filter the data with regard to format (e.g., line breaks and punctuation errors) and content (e.g., data timeliness and hallucination issues).

\paragraph{Training}
During the training process, we fine-tuned a base LLM on the constructed dialogue corpus to obtain the chat LLM:
\begin{equation}
\textbf{LLM}_{chat} = \textit{SFT} \left ( \textbf{LLM}_{base}, ~ \mathcal{D}_{dialogue} \right )
\end{equation}
where $\textbf{LLM}_{base}$ is a pre-trained LLM, $\textbf{LLM}_{chat}$ is the fin-tuned chat model, $\mathcal{D}_{dialogue}$ is the constructed dialogue corpus.

\begin{figure}[t]
  \begin{center}
    \includegraphics[width=\columnwidth]{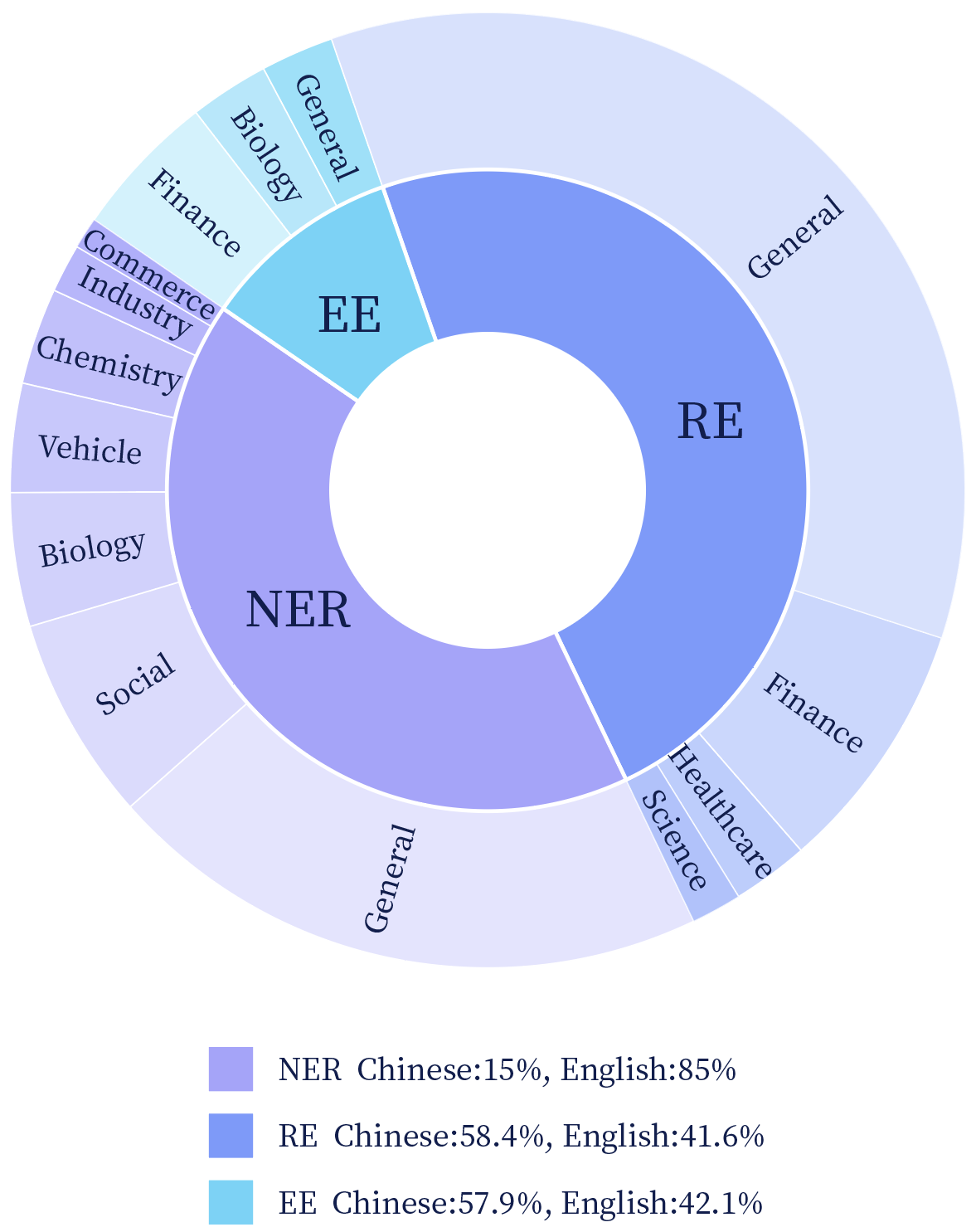}
  \end{center}
  \caption{The distribution of information extraction data.}
  \label{fig-ie-dis}
\end{figure}

\subsection{Instruction Tuning for IE}
After training on the chat data, the chat model gains a fundamental understanding of open-world language and has been further enhanced in its Chinese language capabilities. In the second step of the proposed two-step instruction tuning, we adapt the model to the IE tasks via the IE instruction datasets and standardize the output format of the model. Moreover, we construct the most comprehensive Chinese IE instruction benchmark dataset to support the supervised fine-tuning for the IE tasks. 

\paragraph{Information Extraction Data} 
Due to the lack of Chinese datasets in existing IE benchmarks, we collect 16 Chinese datasets for NER, RE, and EE tasks from diverse domains to build a comprehensive Chinese instruction benchmark, and then combine it with the existing English benchmark IE INSTRUCTIONS \cite{wang2023instructuie}. 

Figure \ref{fig-ie-dis} gives an overview of the English and Chinese IE data for instruction tuning, which includes the distribution of the data across different tasks, domains, and languages. It covers more than 10 domains, such as general, finance, biology, and healthcare. Specifically, the built IE data covers various label types. 

\paragraph{Training}
To enhance the generalization ability, we perform negative sampling on the labels of each instance during the training phrase. For input text $t$ containing $n$ types of labels $L = \{ l_1, l_2, \cdots , l_n \}$, we randomly add several labels to $L$ that do not belong to $L$. During the training process, we fine-tuned the chat LLM on the IE corpus to obtain the universal IE model:
\begin{equation}
\textbf{LLM}_{ie} = \textit{SFT} \left ( \textbf{LLM}_{chat}, ~ \mathcal{D}_{ie} \right )
\end{equation}
where $\textbf{LLM}_{ie}$ is the fine-tuned universal information extraction model, $\mathcal{D}_{ie}$ is the information extraction corpus.

\section{Experiments}
In this section, we conduct experiments under both supervised settings and zero-shot settings to evaluate the effectiveness of YAYI-UIE. For implementation, we choose Baichuan2-13B \cite{yang2023baichuan} as the backbone model and perform the proposed chat-enhanced instruction tuning on it with the $10^{-5}$ learning rate. For the evaluation metrics, we adopt the F1 value to evaluate each dataset in NER, RE and EE tasks in a strict matching manner, and report the respective average F1 of the English datasets and the Chinese datasets on the three tasks. 

\subsection{Experiments on Supervised Settings}

\subsubsection{Datasets}
We conduct supervised experiments on 32 English datasets and 8 Chinese datasets. The English data provided from the benchmark dataset IE INSTRUCTIONS \cite{wang2023instructuie}. Based on IE INSTRUCTIONS, we further collect 8 Chinese datasets to verify the IE capabilities in Chinese under supervised settings. Specifically, for NER task, we adopt CCKS2017 \cite{xia2017clinical}, CCKS2018 \cite{LuoLLYL18}, MSRA \cite{levow-2006-third}, and eCommerce \cite{ecommerce} dataset. For RE task, we adopt DuIE \cite{li2019duie} and InstructIE \cite{gui2023instructie} dataset. For EE task, we adopt DuEE-Fin \cite{han2022duee} DuEE-1.0 \cite{li2020duee}. These datasets cover multiple domains, such as healthcare, finance and biology.

\begin{table}[t]
    \centering
    \setlength{\tabcolsep}{0pt}
    \begin{tabular}{l|ccc|c} 
    \toprule
        Dataset	&	BERT-base	&	UIE  & InstructUIE & YAYI-UIE \\
        \hline
        ACE2005	&	\textbf{87.30}&	85.78 &	86.66&81.78\\
        AnatEM	&	85.82	&	\underline{77.68} &	\textbf{90.89}&76.54\\
        bc2gm	&	80.90&	\underline{74.77} &	\textbf{85.16}&82.05\\
        bc4chemd	&	86.72	&	\underline{82.79} &	\textbf{90.30}&88.46\\
        bc5cdr	&	85.28	&	\underline{78.82} &	\textbf{89.59}&83.67\\
        broadtwitter	&	58.61	&	\underline{67.02} &83.14&\textbf{83.52}\\
        CoNLL03	&	92.40&	92.99&	92.94&\textbf{96.77}\\
        FabNER	&	64.20&	\underline{73.71} &	\textbf{76.20}&72.63\\
        FindVehicle	&	87.13	&	\underline{91.56} &	89.47&	\textbf{98.47}\\
        GENIA-Ent	&	73.30&	\underline{67.46} &	74.71&\textbf{75.21}\\
        HarveyNER	&	82.26	&	\underline{58.13} &	\textbf{88.79}&69.57\\
        MIT Movie	&	88.78	&	\underline{79.56} &	\textbf{89.01}&70.14\\
        MIT Rest.	&	81.02	&	\underline{81.67} &	\textbf{82.55}&79.38\\
        multiNERD	&	91.25	&	\underline{91.75} &	\textbf{92.32}&88.42\\
        ncbi-disease	&	80.20&	\underline{80.13} &	\textbf{90.23}&87.29\\
        Ontonotes	&	\textbf{91.11}	&	\underline{86.25} &	90.19&87.04\\
        polyglot	&	\textbf{75.65}	&	\underline{68.01} &	70.15&70.85\\
        tweetNER7	&	56.49	&	\underline{63.81} &	64.97&\textbf{66.99}\\
        wikiann	&	70.60&	\underline{82.11} &	\textbf{85.13}&72.63\\
        wikineural	&	82.78	&	\underline{\textbf{92.14}} &	91.36&87.63\\
        \hline
        Avg	&	80.09&	78.81  &	\textbf{85.19}&	80.95\\
    \bottomrule
    \end{tabular}
    \caption{ Overall results of YAYI-UIE on English NER datasets. To provide a comprehensive comparison, we conduct experiments on 18 datasets to obtain the experimental results of UIE, which are marked with underlines. }
    \label{tab:ner_en}
\end{table}

\begin{table}[t]
    \centering
    \setlength{\tabcolsep}{11pt}
    \begin{tabular}{l|c|c} 
        \toprule
        Dataset	&	BERT-base  &  YAYI-UIE \\
        \hline
        CCKS 2017 & \textbf{92.68} & 90.73\\
        CCKS 2018 & \textbf{90.82} & 90.39\\
        MSRA &\textbf{ 96.72}& 95.57\\
        eCommerce & 73.70  & \textbf{88.07}\\
        \hline
        Avg	&	88.48& \textbf{91.19}\\
    \bottomrule
    \end{tabular}
    \caption{Overall results of YAYI-UIE on Chinese NER datasets.}
    \label{tab:ner_ch}
\end{table}

\subsubsection{Baselines}
We choose the following representative method as the baselines:
\begin{itemize}
  \item  \textbf{UIE} \cite{lu2022unified} is a unified text-to-structure generation framework that  generates target extraction via schema-based prompts.
  \item  \textbf{USM} \cite{lou2023universal}  is a unified IE tasks framework, which converts IE tasks to a semantic matching problem.
  \item  \textbf{InstructUIE} \cite{wang2023instructuie} proposes a unified information extraction framework based on multi-task instruction tuning.
  \item \textbf{BERT-base} \cite{kenton2019bert} refers to task-specific supervised models with state-of-the-art results based on the pre-trained language model BERT.
\end{itemize}

\subsubsection{Results}

\paragraph{Named Entity Recognition} Table \ref{tab:ner_en} gives the experimental results of the comparative methods and YAYI-UIE on 20 English NER datasets. As shown in the table, YAYI-UIE achieves a higher average F1 value than UIE and BERT-base methods.
When compared to the strong baseline InstructUIE, YAYI-UIE performs slightly worse. The possible reason is the backbone model and training data of InstructUIE are both limited to English only, while the backbone and fine-tuning data of YAYI-UIE are primarily in Chinese, which may affect its capability on English datasets.

Table \ref{tab:ner_ch} gives the experimental results of the comparative methods and YAYI-UIE on 4 Chinese NER datasets. As the existing universal information extraction methods only support the English language, we compare YAYI-UIE to the strong BERT-based methods. In the table, YAYI-UIE achieves the highest average F1 value of 91.19\%. For the CCKS 2017, CCKS 2018 and MSRA, YAYI-UIE is only off by less than 2\% in F1 values, while for eCommerce, it achieves an improvement of 14.37\%. The experimental results show that our model outperforms the baselines on the Chinese NER task. 

\begin{table}[t] 
    \centering
    \setlength{\tabcolsep}{2pt}
    \begin{tabular}{l|ccc|c} 
        \toprule
        Dataset	&	UIE&  USM &InstructUIE &	YAYI-UIE\\
        \hline
        ADE corpus	&	 -	&	 - & 82.31	&	\textbf{84.14}\\
        CoNLL04	    &	75.00&	 78.84 &78.48	&	\textbf{79.73}\\
        GIDS	   &	 -	&	 - &\textbf{81.98}	&	72.36\\
        kbp37	  &	 -	&	 - &36.14	&	\textbf{59.35}\\
        NYT	    &	 -	&	 - &\textbf{90.47	}&	89.97\\
        NYT11 HRL	&	 -	&	 - &56.06	&	\textbf{57.53}\\
        SciERC	&	36.53	&	 37.36 &\textbf{45.15}	&	40.94\\
        semval RE&	 -	&	 - &\textbf{73.23}&	61.02\\
        \hline
        Avg	&	- &	-   &	67.98	& \textbf{68.13}\\
    \bottomrule
    \end{tabular}
    \caption{Overall results on English RE datasets.}
    \label{tab:re_en}
\end{table}

\begin{table}
    \centering
    \setlength{\tabcolsep}{13pt}
    \begin{tabular}{l|c|c} 
        \toprule
        Dataset	&	 BERT-base&  YAYI-UIE \\
        \hline
        DuIE	&	 74.30&	\textbf{81.19}\\
        InstructIE	&	 49.21&	\textbf{59.52}\\
        \hline
        Avg	&	61.76&	\textbf{70.36}\\
        \bottomrule
    \end{tabular}
    \caption{Overall results on Chinese RE datasets.}
    \label{tab:re_ch}
\end{table}

\begin{table*}[t]
    \centering
    \setlength{\tabcolsep}{10pt}
    \begin{tabular}{l|l|cccc|c} 
        \toprule
        ~ & Dataset	&	BERT-base&  USM  &  UIE& InstructUIE &	YAYI-UIE\\
        \hline
        \multirow{4}*{Event Trigger} & ACE2005	&	72.5& 72.41 & 73.36&  \textbf{77.13} & 65.00\\
         ~&   CASIE	&	68.98& \textbf{71.73} & 69.33&  67.80  &   63.00\\
         ~&   PHEE	&	- & - & \underline{64.77}   &  \textbf{70.14}  &   63.00\\
         \cline{2-7}
         ~&   Avg	&	-& - & 69.15&  \textbf{71.69}  &   63.67\\
        \hline
        \multirow{4}*{Event Argument} & ACE2005	  &	59.9& 55.83  & 54.79&  \textbf{72.94} &     62.71\\
         ~&   CASIE                               & 60.37& 63.26  & 61.3&   63.53  &   \textbf{64.23}\\
         ~&   PHEE                               &	-     & -     & \underline{63.70}   &   62.91   &   \textbf{77.19}\\
         \cline{2-7}
         ~&   Avg                               &	-     & -    & 59.93&  66.46    &   \textbf{68.04}\\
        \bottomrule
    \end{tabular}
    \caption{ Overall results on English EE datasets. The results marked with underlines are reproduced in this paper.}
    \label{tab:ee_en}
\end{table*}

\paragraph{Relation Extraction} Table \ref{tab:re_en} gives the experimental results of the comparative models and YAYI-UIE on 8 English RE datasets. We can see from the table that YAYI-UIE achieves the highest average F1 value, and gains a significant improvement on kbp37 compared to the strong baseline InstructUIE. Compared with UIE and USM, YAYI-UIE performs better on the 2 datasets. It should be noted that the model and code for USM are not available, and we cannot reproduce UIE on these RE datasets due to the lack of position information. 

Table \ref{tab:re_ch} gives the experimental results of the comparative models and YAYI-UIE on 2 Chinese RE datasets. From the table, we can see that YAYI-UIE achieves the highest average F1 value, and gains 8.6\% F1  improvement compared to the baselines. In general, the experimental results demonstrate the effectiveness of our YAYI-UIE for both English and Chinese RE tasks.

\paragraph{Event Extraction} Table \ref{tab:ee_en} gives the Event Trigger and Event Argument F1 value experimental results of the comparative models and YAYI-UIE on 3 English EE datasets. Our YAYI-UIE achieves the highest average F1 score for the event argument extraction task. Compared with UIE, YAYI-UIE performs better on 3 out of 6 datasets, while compared with InstructUIE, YAYI-UIE performs better on 2 out of 6 datasets on EE task. 

Table \ref{tab:ee_ch} gives the experimental results of the comparative models and YAYI-UIE on 2 Chinese EE datasets. In the table, we can see that YAYI-UIE achieves the highest average F1 score for both the event trigger and argument extraction tasks. The experimental results demonstrate that our model outperforms the comparative models on the Chinese EE task. 

\begin{table} [t]
    \centering
    \setlength{\tabcolsep}{3pt}
    \begin{tabular}{l|l|c|c} 
        \toprule
        ~ & Dataset	&	UIE  &	YAYI-UIE\\
        \hline
        \multirow{3}*{Event Trigger} & DuEE-Fin	& \textbf{84.53}& 82.50  \\
        ~&   DuEE-1.0   & 82.18& \textbf{85.00}	\\
        \cline{2-4}
         ~&   Avg     &  83.36& \textbf{83.75}    \\
        \hline
        \multirow{3}*{Event Argument} & DuEE-Fin & \textbf{75.73}& 70.02  \\
        ~&   DuEE-1.0   & 70.68& \textbf{78.08 }\\
        \cline{2-4}
         ~&   Avg     &  73.21& \textbf{74.05}  \\
        \bottomrule
    \end{tabular}
    \caption{Overall results on Chinese EE datasets.}
    \label{tab:ee_ch}
\end{table}

\subsection{Experiments on Zero-shot Settings}


\subsubsection{Datasets}

To validate the zero-shot capability of YAYI-UIE, we collected 16 datasets and tested their performance on three tasks separately, which do not appear in the training set. For NER task, we evaluate English capability on five CrossNER \cite{liu2021crossner} subsets (AI, literature, music, politics, science) and Chinese capability on the datasets of boson \footnote{\url{https://github.com/InsaneLife/ChineseNLPCorpus/tree/master/NER/boson}}, clue \cite{xu-etal-2020-clue} and weibo \cite{peng-dredze-2015-named}. For RE task, we evaluate the model's English ability on FewRel \cite{han-etal-2018-fewrel} and Wiki-ZSL \cite{chen-li-2021-zs}, and Chinese capability on SKE 2020 \footnote{\url{https://aistudio.baidu.com/datasetdetail/177191}}, COAE 2016 \footnote{\url{https://github.com/Sewens/COAE2016}}, IPRE \cite{wang2019ipre}. For EE task, we test the event argument and event trigger extraction separately, which use Commodity News Corpus \cite{lee-etal-2022-crudeoilnews} for the English capability, FewFC \cite{zhou2021role} and CCF law \footnote{\url{https://aistudio.baidu.com/projectdetail/4201483}} for the Chinese capability.

\begin{table*}[t]
    \centering
    \setlength{\tabcolsep}{4.4pt}
    \begin{tabular}{l|ccccc|c|ccc|c} 
    \toprule
        \multirow{2}*{Method} & \multicolumn{6}{c|}{EN} & \multicolumn{4}{c}{CH}  \\
        \cline{2-11}
         ~ &	AI	&Literature	& Music	& Politics	& Science & Avg 	&boson	&clue	& weibo & Avg\\
        \hline
        
        ChatGPT & \textbf{54.40} & 54.07 & \textbf{61.24} & \textbf{59.12} & \textbf{63.00} & \textbf{58.37} &  38.53 & 25.44 & 29.3 & 31.09 \\
        ChatGLM2-6b & 0.01 & 0.03 & 0.00 & 0.46 & 0.68	& 0.24 &  1.13 & 0.07 & 8.09 & 3.10\\
        UIE & 31.14 & 38.97 & 33.91 & 46.28 & 41.56 & 38.37 &  40.64 & 34.91 & \textbf{40.79} & 38.78 \\
        USM & 28.18	& \textbf{56.00} & 44.93 & 36.10 & 44.09	& 41.86 &  - & - & - & - \\
        InstructUIE & 49.00 & 47.21 & 53.16 &	48.15 &	49.30 & 49.36 &  - & - & - &  - \\ 
        KnowLM & 13.76 & 20.18 & 14.78 & 33.86 & 9.19 & 18.35 &  25.96 &4.44  & 25.20 &  18.53 \\
        \hline
        YAYI-UIE & 52.40& 45.99& 51.20& 51.82& 50.53& 50.39&  \textbf{49.25}& \textbf{36.46}& 36.78&	\textbf{40.83}\\
        \bottomrule
    \end{tabular}
    \caption{ Zero-shot performance on NER task, including 5 English datasets and 3 Chinese datasets.}
    \label{tab:0-shot-ner}
\end{table*}

\begin{table*} 
    \centering
    \setlength{\tabcolsep}{8pt}
    \begin{tabular}{l|cc|c|ccc|c} 
    \toprule
    \multirow{2}*{Method}& \multicolumn{3}{c|}{EN} & \multicolumn{4}{c}{CH} \\
        \cline{2-8} 
         &	FewRel	& Wiki-ZSL	& Avg & SKE 2020	& COAE2016	& IPRE & Avg \\
        \hline 
        gpt-3.5-turbo & 9.96 & 13.14 & 11.55 & 24.47 & 19.31 & 6.73 & 16.84\\
        ZETT(T5-small) & 30.53 & 31.74 & 31.14 & - & - & - & - \\
        ZETT(T5-base) & 33.71 & 31.17 & 32.44 & - & - & - & - \\
        InstructUIE & \textbf{39.55} & 35.20	& 37.38	& -	& - & - & -\\ 
        KnowLM & 17.46 & 15.33 & 16.40 & 0.40& 6.56 & 9.75 & 5.57 \\
        \hline
        YAYI-UIE	& 36.09& \textbf{41.07}& \textbf{38.58}&\textbf{ 70.8}& \textbf{19.97}& \textbf{22.97}& \textbf{37.91}\\
        \bottomrule
    \end{tabular}
    \caption{Zero-shot performance on RE task, including 2 English datasets and 3 Chinese datasets.}
    \label{tab:0-shot-re}
\end{table*}

\begin{table*}
    \centering
    \setlength{\tabcolsep}{13pt}
    \begin{tabular}{c|c|c|cc|c} 
    \toprule
    
        \multirow{2}*{~} &\multirow{2}*{Method} & \multicolumn{1}{c|}{EN} & \multicolumn{3}{c}{CH}  \\
        \cline{3-6}
        ~ &~ &	commodity news & FewFC & CCF law & Avg \\
        \hline
        \multirow{4}*{Event Trigger}   &ChatGPT & 1.41 & 16.15 & 0.00 & 8.08\\
        &UIE & - & 50.23 & 2.16 & 26.20 \\
        &InstructUIE & \textbf{23.26} & - & - & - \\
        \cline{2-6}
        &YAYI-UIE & 12.45&\textbf{ 81.28}& \textbf{12.87}& \textbf{47.08}\\
    
        \hline
        
         \multirow{4}*{Event Argument}   &ChatGPT & 8.60 & 44.40 & 44.57 & 44.49\\
        ~ &UIE & - & 43.02 & \textbf{60.85} & 51.94\\
        ~& InstructUIE & \textbf{21.78} & - & -& - \\
        \cline{2-6}
        ~& YAYI-UIE &	19.74& \textbf{63.06}& 59.42& \textbf{61.24}\\
        
        \bottomrule 
    \end{tabular}
    \caption{ Zero-shot performance on EE task, including 1 English datasets and 2 Chinese datasets.}
    \label{tab:0-shot-ee}
\end{table*}

\begin{table*} 
    \centering
    \setlength{\tabcolsep}{12pt}
    \begin{tabular}{l|cc|cc} 
    \toprule
        \multirow{2}*{Task} & \multicolumn{2}{c|}{EN} & \multicolumn{2}{c}{CH} \\
        \cline{2-5}
         ~ &	Baichuan-base & Baichuan-chat & Baichuan-base & Baichuan-chat \\
        \hline
        NER	& 67.21 & \textbf{81.21} & 66.98 &\textbf{89.15} \\
        RE & 48.99 & \textbf{65.78} & 44.45 & \textbf{62.67} \\
        Event Argument & 44.44 & \textbf{63.48} & 63.03 & \textbf{68.98} \\
        Event Trigger & 45.67 & \textbf{61.33} & 74.31 & \textbf{84.50} \\
        \hline
        Avg & 51.58  & \textbf{67.95} & 62.19 & \textbf{76.34} \\
        \bottomrule
    \end{tabular}
    \caption{Baichuan-13B-chat and Baichuan-13B-base's performance on each IE task}
    \label{ablations_study}
\end{table*}

\subsubsection{Baselines}
We choose the following representative models for comparative baselines:
\begin{itemize}
    \item  \textbf{ZETT} \cite{kim2022zero} is a framework that extracts relation triplets from unstructured text. 
    \item  \textbf{ChatGPT} \cite{ouyang2022training} is a state-of-the-art conversational AI language model that is built upon the GPT-3.5 architecture.
    \item \textbf{ChatGLM} \cite{du2022glm} is an open-source, Chinese-English bilingual conversation language model.
    \item \textbf{KnowLM} \cite{knowlm} is an open-source and extensible knowledge graph extraction tool that can extract entities and relations.
\end{itemize}

\subsubsection{Results}

\paragraph{Named Entity Recognition} Table \ref{tab:0-shot-ner} gives the zero-shot experimental results of the comparative models and YAYI-UIE on 5 unseen English NER datasets and 3 unseen Chinese NER datasets. For the English NER task, YAYI-UIE outperforms several strong baselines except for ChatGPT. For the Chinese NER task, YAYI-UIE achieves the highest average F1 score.

\paragraph{Relation Extraction} Table \ref{tab:0-shot-re} gives the zero-shot experimental results of the comparative models and YAYI-UIE on 2 unseen English RE datasets, and 3 unseen Chinese RE datasets. We can observe that YAYI-UIE achieves the SOTA on both English and Chinese datasets. For the English RE task, YAYI-UIE outperforms the best comparative model InstructUIE in average F1 score by 1.2\%. The performance on FewRel is not obviously due to the small size and insufficient learning of the model. For the Chinese RE task, our proposed model performs much better than the baselines. 

\paragraph{Event Extraction} Table \ref{tab:0-shot-ee} gives the zero-shot experimental results of the comparative models and YAYI-UIE on 1 unseen English EE dataset and 2 unseen Chinese EE datasets. The result shows that YAYI-UIE achieves the SOTA performance for Chinese EE task, and also
comparable performance for English EE task. It is worth mentioning that InstructUIE only has English capability, and our model has added a large amount of Chinese data in the training, which has reduced the English capability of the model to some extent. 

\section{Ablation Study}
We conduct the ablation study to further evaluate the effectiveness of the instruction tuning for chat using dialogue data in our framework. We conduct separate experiments with Baichuan2-13B-base and Baichuan2-13B-chat \cite{yang2023baichuan} as the backbone model for IE instruction fine-tuning. The performances of the two models are measured by calculating the strict F1 score for each dataset. We report the average F1 score for each task. Table \ref{ablations_study} shows that Baichuan2-chat's performance for each task significantly outperforms Baichuan2-13B-chat by more than 10 points, which verifies the effectiveness of the chat fine-tuning for the universal information extraction task.

\section{Related Work}
\paragraph{Large Language Models} The advent of large language models (LLMs) has instigated a revolutionary paradigm shift within the field of natural language processing \cite{guo2023close,qin2023chatgpt,bubeck2023sparks}. 
LLMs, such as LLaMA \cite{llama,llama2}, ChatGPT \cite{ouyang2022training} and GPT4 \cite{gpt4}, have exhibited remarkable abilities across various applications. 
These LLMs undergo three primary training stages: pre-training, supervised fine-tuning (SFT), and reinforcement learning from human feedback (RLHF). 
During the pre-training phase, LLMs gain extensive skills and knowledge. However, they face a challenge in adhering to specific instructions. To mitigate this limitation, SFT is incorporated as a supplementary step. This process entails additional training of the LLM utilizing a dedicated annotated dataset that includes instructions and corresponding responses, augmenting its capabilities in accurately following instructions. 
RLHF, by incorporating human feedback into the training loop, serves as a pivotal mechanism for steering LLMs toward generating high-quality and harmless responses. In this paper, we propose a chat-enhanced instruction tuning framework for universal information extraction, which consists of two steps in the SFT process. We first utilize dialogue data to fine-tune a base LLM for obtaining instructions following abilities, and then fine-tune the chat model to generate the target IE structure.

\paragraph{Information Extraction} Information extraction constitutes a longstanding field devoted to the automated extraction of diverse information structures from unstructured textual sources. 
Classic IE methods \cite{mengge-etal-2020-coarse,wang-etal-2021-cleve,qin-etal-2021-erica,wang-etal-2022-miner} necessitate the formulation of task-specific architectures and the training of dedicated models, which reveals limitations in the generalization ability of models across diverse IE tasks and imposes stringent demands for annotated data. 
Lu et al. \cite{lu2022unified} have recently pioneered UIE by uniformly modeling IE tasks with a text-to-structure framework. However, a notable limitation of UIE lies in its deficiency in transferring learning capabilities across diverse tasks or schemas. 
Lou et al. \cite{lou2023universal} proposed USM by designing three directed token-linking operations to decouple task-specific IE tasks into two extraction abilities, resulting in a notable increase in both training and inference time.
Wang et al. \cite{wang2023instructuie} proposed InstructUIE by utilizing instructive guidance to direct LLMs toward the task, facilitating the generation of target structures. Unfortunately, this method is deficient in IE capabilities for Chinese languages other than English. In this paper, we propose an end-to-end framework YAYI-UIE for universal information extraction that supports both Chinese and English.

\section{Conclusion}
In this paper, we propose a chat-enhanced instruction tuning framework YAYI-UIE for universal information extraction,  and build the most comprehensive Chinese IE instruction benchmark. The proposed framework consists of two instruction-tuning steps. It first utilizes dialogue data to fine-tune a base LLM for obtaining common understanding abilities, and then utilizes the constructed Chinese IE benchmark dataset along with the existing English benchmark for IE instruction fine-tuning. Experimental results show that our proposed framework achieves state-of-the-art performance in Chinese while maintaining English language capabilities. 



\bibliography{main}

\appendix



\end{document}